%% file: main.tex
\documentclass{article} 
\usepackage{aaai24_arxiv}  
\usepackage{times}  
\usepackage{helvet}  
\usepackage{courier}  
\usepackage[hyphens]{url}  
\usepackage{graphicx} 
\usepackage{natbib}  
\usepackage{caption} 
\usepackage{algorithm}
\usepackage{algorithmic}

\usepackage{epsfig}
\usepackage{graphicx}
\usepackage{amsmath}
\usepackage{amssymb}

\usepackage{url}
\usepackage{caption}
\usepackage{xspace}
\usepackage{array}
\usepackage{makecell}
\usepackage{comment}
\usepackage{subfiles}
\usepackage{enumitem}
\usepackage{multirow}
\usepackage{makecell}
\usepackage{xcolor,colortbl}
\definecolor{citecolor}{HTML}{0071BC}
\definecolor{linkcolor}{HTML}{ED1C24}

\usepackage[pagebackref=false, breaklinks=true, colorlinks, citecolor=citecolor, linkcolor=linkcolor, bookmarks=false]{hyperref}
\usepackage{pifont}
\newcommand{\tline}{\Xhline{1pt}}
\newcommand{\envelope}{\ding{41}}

\newcommand{\apcoco}{AP & AP$_{\text{50}}$ & AP$_{\text{75}}$ & AP$_{\text{S}}$ & AP$_{\text{M}}$ & AP$_{\text{L}}$}

\newcommand{\aps}{AP & AP$_{\text{50}}$ & AP$_{\text{75}}$ }
\newcommand{\apscoco}{AP & AP$_{\text{50}}$ & AP$_{\text{75}}$ }

\newcommand{\etal}{\textit{et al}.}
\newcommand{\ie}{\textit{i}.\textit{e}.}
\newcommand{\eg}{\textit{e}.\textit{g}.}
\newcommand{\tb}[1]{\textbf{#1}}

\makeatletter\renewcommand\paragraph{\@startsection{paragraph}{4}{\z@}
  {.5em \@plus1ex \@minus.2ex}{-.5em}{\normalfont\normalsize\bfseries}}\makeatother

\usepackage{newfloat}
\usepackage{listings}
\DeclareCaptionStyle{ruled}{labelfont=normalfont,labelsep=colon,strut=off} 
\floatstyle{ruled}
\newfloat{listing}{tb}{lst}{}
\floatname{listing}{Listing}
\newcommand{\name}{MobileInst}

\title{\name{}: Video Instance Segmentation on the Mobile}
\author{
    Renhong Zhang$^{1 *}$,~
    Tianheng Cheng$^{1 *}$,~
    Shusheng Yang$^{1}$,~
    Haoyi Jiang$^{1}$,~
    Shuai Zhang$^2$,\\
    Jiancheng Lyu$^2$,~ 
    Xin Li$^2$,~
    Xiaowen Ying$^2$,~
    Dashan Gao$^2$,~
    Wenyu Liu$^1$,~
    Xinggang Wang$^{1}$\textsuperscript{\envelope}
}
\affiliations{
$^1$~School of EIC, Huazhong University of Science \& Technology~
$^2$~Qualcomm AI Research$^\dag$
}

\usepackage{bibentry}

\begin{document}

\maketitle

\let\thefootnote\relax\footnotetext{$^*$ These authors contributed equally. \textsuperscript{\envelope} Corresponding author: Xinggang Wang (\url{xgwang@hust.edu.cn}). $^\dag$ Qualcomm AI Research is an initiative of Qualcomm Technologies, Inc. $^\ddag$ Snapdragon and Qualcomm branded products are products of Qualcomm Technologies, Inc. and/or its subsidiaries.}

\begin{abstract}
Video instance segmentation on mobile devices is an important yet very challenging  edge AI problem. It mainly suffers from (1) heavy computation and memory costs for frame-by-frame pixel-level instance perception and (2) complicated heuristics for tracking objects.
To address those issues, we present \textbf{\name{}}, a lightweight and mobile-friendly framework for video instance segmentation on mobile devices.
Firstly, \name{} adopts a mobile vision transformer to extract multi-level semantic features and presents an efficient query-based dual-transformer instance decoder for mask kernels and a semantic-enhanced mask decoder to generate instance segmentation per frame.
Secondly, \name{} exploits simple yet effective kernel reuse and kernel association to track objects for video instance segmentation.
Further, we propose temporal query passing to enhance the tracking ability for kernels.
We conduct experiments on COCO and YouTube-VIS datasets to demonstrate the superiority of \name{} and evaluate the inference latency on one single CPU core of Snapdragon$^\circledR$ 778G Mobile Platform\footnote{Snapdragon and Qualcomm branded products are products of Qualcomm Technologies, Inc. and/or its subsidiaries.}, without other methods of acceleration.
On the COCO dataset, \name{} achieves 31.2 mask AP and 433 ms on the mobile CPU, which reduces the latency by 50\% compared to the previous SOTA.
For video instance segmentation, \name{} achieves 35.0 AP on YouTube-VIS 2019 and 30.1 AP on YouTube-VIS 2021.
Code will be available to facilitate real-world applications and future research.

\end{abstract}

\section{Introduction}
\label{sec:intro}
\subfile{tex/intro.tex}
\section{Related Work}
\label{sec:relatedw}

\subfile{tex/related.tex}
\section{\name{}}
\label{sec:method}
\subfile{tex/method.tex}
\section{Experiments}
\label{sec:experiments}

\subfile{tex/exp.tex}

\section{Visualizations}
\label{sec:vis}
\subfile{tex/vis.tex}

\section{Conclusion}
In this paper, we propose \name{}, an elaborate-designed video instance segmentation framework for mobile devices.
To reduce computation overhead, we propose an efficient query-based dual-transformer instance decoder and a semantic-enhanced mask decoder, with which \name{} achieves competitive performance and maintains a satisfactory inference speed simultaneously.
We also propose an efficient method to extend our \name{} to video instance segmentation tasks without introducing extra parameters.
Experimental results on both COCO and Youtube-VIS datasets demonstrate the superiority of \name{} in terms of both accuracy and inference speed.
We hope our work can facilitate further research on instance-level visual recognition on resource-constrained devices.

\paragraph{Acknowledgement.}
This work was partially supported by the National Key Research and Development Program of China under Grant 2022YFB4500602, the National Natural Science Foundation of China (No. 62276108), and the University Research Collaboration Project (HUA-474829) from Qualcomm.

\bibliography{aaai24}

\end{document}

%% file: tex/intro.tex
Deep visual understanding algorithms with powerful GPUs have achieved great success, but their performance is reaching a plateau. Edge AI, which enables massive low-resource computing devices, is becoming increasingly popular. In this paper, we study a very challenging edge AI task, namely video instance segmentation (VIS) on mobile devices.
The goal of VIS \cite{VIS} is to simultaneously identify, segment, and track objects in the video sequence and it attracts a wide range of applications, \eg, robotics, autonomous vehicles, video editing, and augmented reality.
The advances in deep convolutional neural networks and vision transformers have made great progress in video instance segmentation and achieved tremendous performance \cite{MaskPropagation, STEm-Seg, seqmaskrcnn} on GPUs.
Nevertheless, many real-world applications tend to require those VIS methods to run on resource-constrained devices, \eg, mobile phones, and inference with low latency.
It's challenging but urgent to develop and deploy efficient approaches for video instance segmentation on mobile or embedded devices.

\begin{figure}
    \centering
    \includegraphics[width=0.95\linewidth]{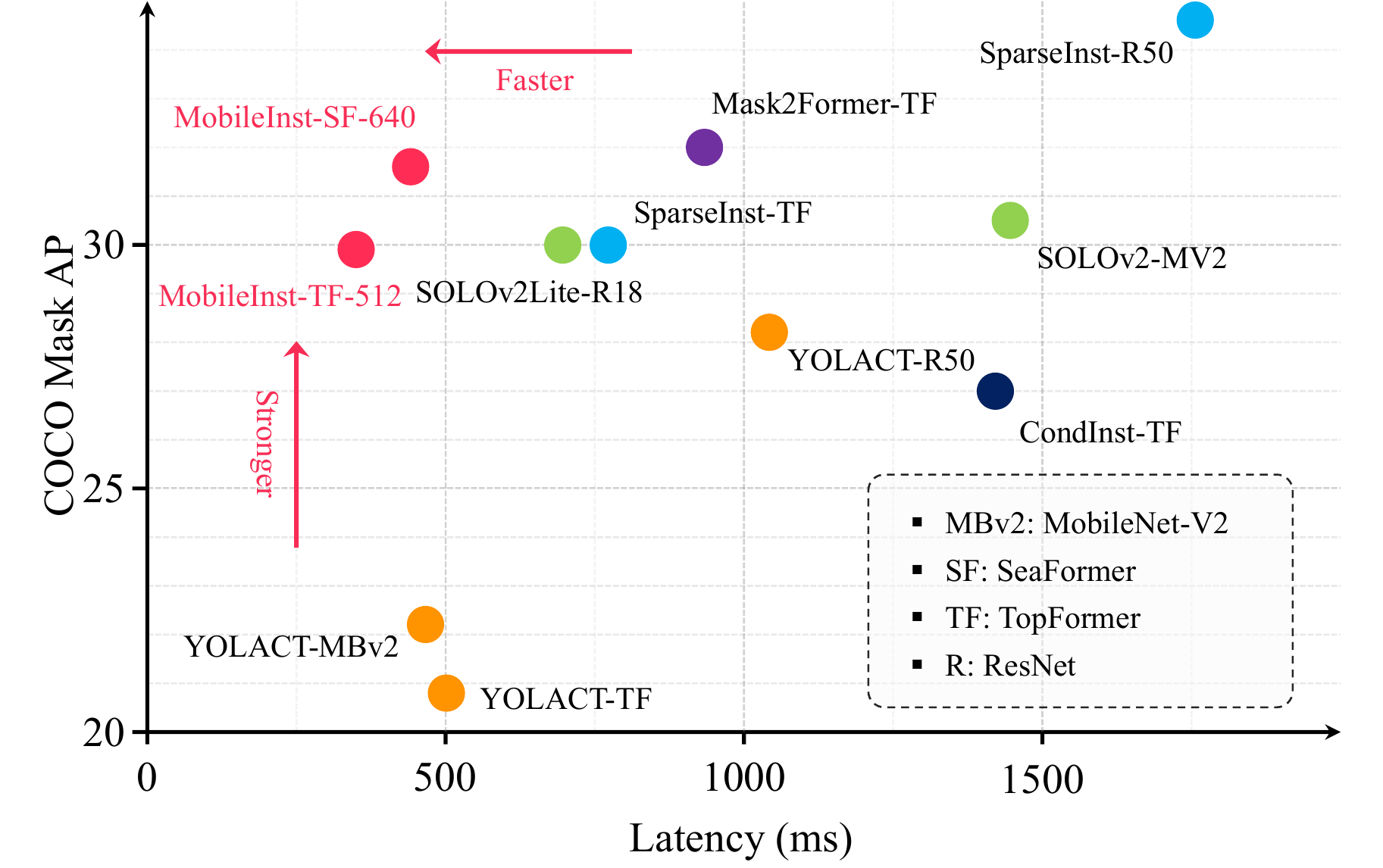}
    \vspace{-10pt}
    \caption{\textbf{Speed-and-Accuracy Trade-off.} We evaluate all models on COCO \texttt{test-dev} and inference speeds are measured on one mobile CPU, \ie, Snapdragon 778G. The proposed \name{} outperforms other methods in both speed and accuracy on mobile devices.}
    \label{fig:speed_and_accuracy}
    \vspace{-15pt}
\end{figure}

Albeit great progress has been witnessed in the VIS field, there are several obstacles that prevent modern VIS frameworks from being deployed on edge devices with limited resources, such as mobile chipsets.
Prevalent methods for video instance segmentation can be categorized into two groups: offline methods (clip-level) and online methods (frame-level).
Offline methods~\cite{vistr,ifc,yang2022tevit,wu2022seqformer,heo2022vita, seqmaskrcnn} divide the video into clips, generate the instance predictions for each clip, and then associate the instances by instance matching across clips.
However, inference with clips (multiple frames) is infeasible in mobile devices in terms of computation and memory cost.
Whereas, online methods~\cite{VIS,Crossvis,SipMask,compfeat,idol} forward and predict with frame-level input but require complicated heuristic procedures to associate instances across frames, \eg, NMS, which are inefficient in mobile devices.
In addition, recent methods for video instance segmentation tend to employ heavy architectures, especially for the methods based on transformers, which incur a large computation burden and memory costs.
Directly scaling down the model size for lower inference latency will inevitably cause severe performance degradation, which limits the practical application of recent methods.
Designing and deploying video instance segmentation techniques for resource-constrained devices have not been well explored yet, which are not trivial but crucial for real-world applications.
\begin{figure*}[h]
    \centering
    \includegraphics[width=\linewidth]{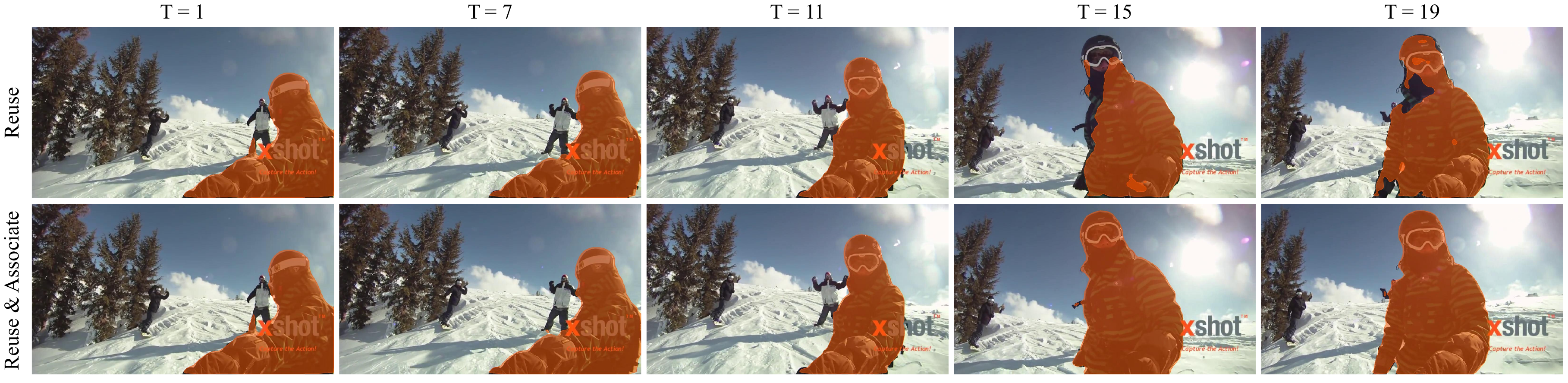}
    \vspace{-20pt}
    \caption{\textbf{Reusing kernels for tracking.} We train \name{} for single-frame instance segmentation on YouTube-VIS 2019, and then apply \name{} to infer the per-frame segmentation and track objects via reusing mask kernels. \textbf{The upper row:} We adopt the predicted mask kernel in $T=1$ frame to obtain the segmentation results in the video sequence.
    In a short time, the reused mask kernels provide accurate segmentation and tracking results.
    \textbf{The bottom row:} We divide the videos into K-frame clips and reuse the mask kernels of every first frame. In addition, we adopt simple yet effective cosine similarity to associate the kernels in consecutive clips ($K$ is set to 3). Reusing kernels with association performs well and is efficient.}
    \label{fig:kernel_reuse}
    \vspace{-15pt}
\end{figure*}

In this paper, we introduce \name{} to achieve performant video instance segmentation on mobile devices for the first time.
\name{} is efficient and mobile-friendly from two key aspects (1) lightweight architectures for segmenting objects per frame and (2) simple yet effective temporal modeling for tracking instances across frames.
Specifically, \name{} consists of a query-based \textit{dual transformer instance decoder}, which exploits object queries to segment objects, updates object queries through global contexts and local details, and then generates the mask kernels and classification scores.
To efficiently aggregate multi-scale features and global contexts for mask features, \name{} employs a \textit{semantic-enhanced mask decoder}.
%
The object queries are forced to represent objects in a one-to-one manner and we discover that mask kernels (generated by object queries) tend to be \textit{temporally consistent} in consecutive frames, \ie, the same kernel (query) corresponds to the same objects in nearby frames, as shown in Fig.~\ref{fig:kernel_reuse}.
Therefore, we exploit simple yet effective \textit{kernel reuse} and \textit{kernel association} to track objects by reusing kernels in a $T$-frame clips and associate objects across clips by kernel cosine similarity.
Further, we present temporal query passing to enhance the tracking ability for object queries during training with video sequences. 
\name{} can one-the-fly segment and track objects in videos on mobile devices.

We evaluate \name{} on the COCO dataset~\cite{COCOLinMBHPRDZ14} and YouTube-VIS~\cite{VIS} for image and video instance segmentation, respectively.

The main contributions of this paper can be summarized as follows:
\begin{itemize}
    \item We present a cutting-edge and mobile-friendly framework named \name{} for video instance segmentation on mobile devices, which is the first work targeting VIS on mobile devices to the best of our knowledge.
    \item We propose a dual transformer instance decoder and semantic-enhanced mask decoder in \name{} for efficiently segmenting objects in frames.
    \item We present kernel reuse and kernel association for tracking objects across frames which are simple and efficient along with the temporal training strategy.
    \item We benchmark the mobile VIS problem by implementing a wide range of lightweight VIS methods for comparisons. The proposed \name{} can achieve state-of-the-art mobile VIS performance, \ie, 35.0 AP with 188 ms on YouTube-VIS-2019 and 31.2 AP with 433 ms on COCO \texttt{test-dev}, when deployed on the CPU of Snapdragon 778G, without using mixed precision, low-bit quantization, or the inside hardware accelerator for neural network inference. 
\end{itemize}

%% file: tex/related.tex
\subsection{Instance Segmentation}
Most methods address instance segmentation by extending object detectors with mask branches for instance masks, \eg, Mask R-CNN~\cite{HeGDG17} adds an RoI-based fully convolutional network upon Faster R-CNN~\cite{faster} to predict object masks.
\cite{TianSC20,YolactBolyaZXL19,PolarMaskXieSSWLLSL20,MEInstZhangTSYY20} present single-stage methods for instance segmentation.
Several methods~\cite{SOLOWangKSJL20,SOLOV2WangZKLS20,ChengWCZZHZ022} present detector-free instance segmentation for simplicity and efficiency.
Recently, query-based detectors~\cite{DETRCarionMSUKZ20,DeDETRZhuSLLWD21,queryinstabs-2105-01928,mask_former_abs-2107-06278,yolos} reformulate object detection with set prediction and show promising results on instance segmentation.
Considering the inference speed, YOLACT~\cite{YolactBolyaZXL19} and SparseInst~\cite{ChengWCZZHZ022} propose real-time approaches and achieve a good trade-off between speed and accuracy.
However, existing methods are still hard to deploy to mobile devices for practical applications due to the large computation burden and complex post-processing procedures.
The proposed \name{} designs efficient transformers and aims to achieve better speed and accuracy on mobile devices.

\subsection{Video Instance Segmentation}

\paragraph{Offline Methods.} 
For offline VIS, the models~\cite{vistr,ifc,yang2022tevit,wu2022seqformer,heo2022vita} take a video clip as the input once, achieving good performance due to the rich contextual information.
VisTR~\cite{vistr} proposes the first fully end-to-end offline VIS method. 
Several works effectively alleviate the computation burden brought by self-attention by building Inter-frame Communication Transformers~\cite{ifc}, using messengers to exchange temporal information in the backbone~\cite{yang2022tevit}, and focusing on temporal interaction of instance between frames~\cite{wu2022seqformer,heo2022vita}. However, clip-level input is difficult to apply to resource-constrained mobile devices.

\paragraph{Online Methods.}
For online VIS, early models~\cite{VIS,Crossvis,visolo} extend CNN-based image segmentation models to handle temporal coherence by employing an extra embedding to identify instances and associating instances with heuristic algorithms.
However, those methods based on dense prediction require extra complex post-processing steps, \eg, NMS,  which hinders end-to-end inference on mobile devices.
Li \etal propose the temporal pyramid routing~\cite{LiHYDYCTT23} for aggregating adjacent frames to improve the efficiency of VIS.
Recently, many excellent transformer-based models address VIS by using simple tracking heuristics with object queries which have capabilities 
of distinguishing instances~\cite{minvis}. IDOL~\cite{idol} obtains performance comparable to offline VIS by contrastive learning of the instance embedding between different frames. InsPro~\cite{inspro} and InstanceFormer~\cite{insformer} respectively use proposals and reference points to establish correspondences between instance information for online temporal propagation. Unfortunately, existing models rely on large-scale models like Mask2Former~\cite{mask2former} and Deformable DETR~\cite{DeDETRZhuSLLWD21} beyond the capabilities of many mobile devices.

\subsection{Mobile Vision Transformers}
Vision transformers (ViT)~\cite{ViT} have demonstrated immense power in various vision tasks. 
Subsequent works~\cite{swintransformer,pvt,msgtransformer} adopt hierarchical architectures and incorporate spatial inductive biases or locality into vision transformers, which provide better feature representation for downstream tasks. 
Vision transformers tend to be resource-consuming compared to convolutional networks due to the multi-head attention~\cite{attention}.
To facilitate the mobile applications, MobileViT~\cite{mobilevit} and TopFormer~\cite{TopFormer} design mobile-friendly transformers by incorporating efficient transformer blocks into MobileNetV2~\cite{Sandler2018MobileNetV2IR}.
Mobile-Former~\cite{Chen2021MobileFormerBM} presents a parallel architecture containing a transformer and a MobileNet~\cite{Sandler2018MobileNetV2IR} and adopts attention to exchange features between convolutional networks and transformers. 
Recently, Wan~\etal propose SeaFormer~\cite{SeaFormer} with axial attention which is more efficient and effective.
In this paper, \name{} aims for video instance segmentation on mobile devices, which is more challenging than designing mobile backbones.

\begin{figure*}[htbp]
    \centering
    \includegraphics[width=0.95\linewidth]{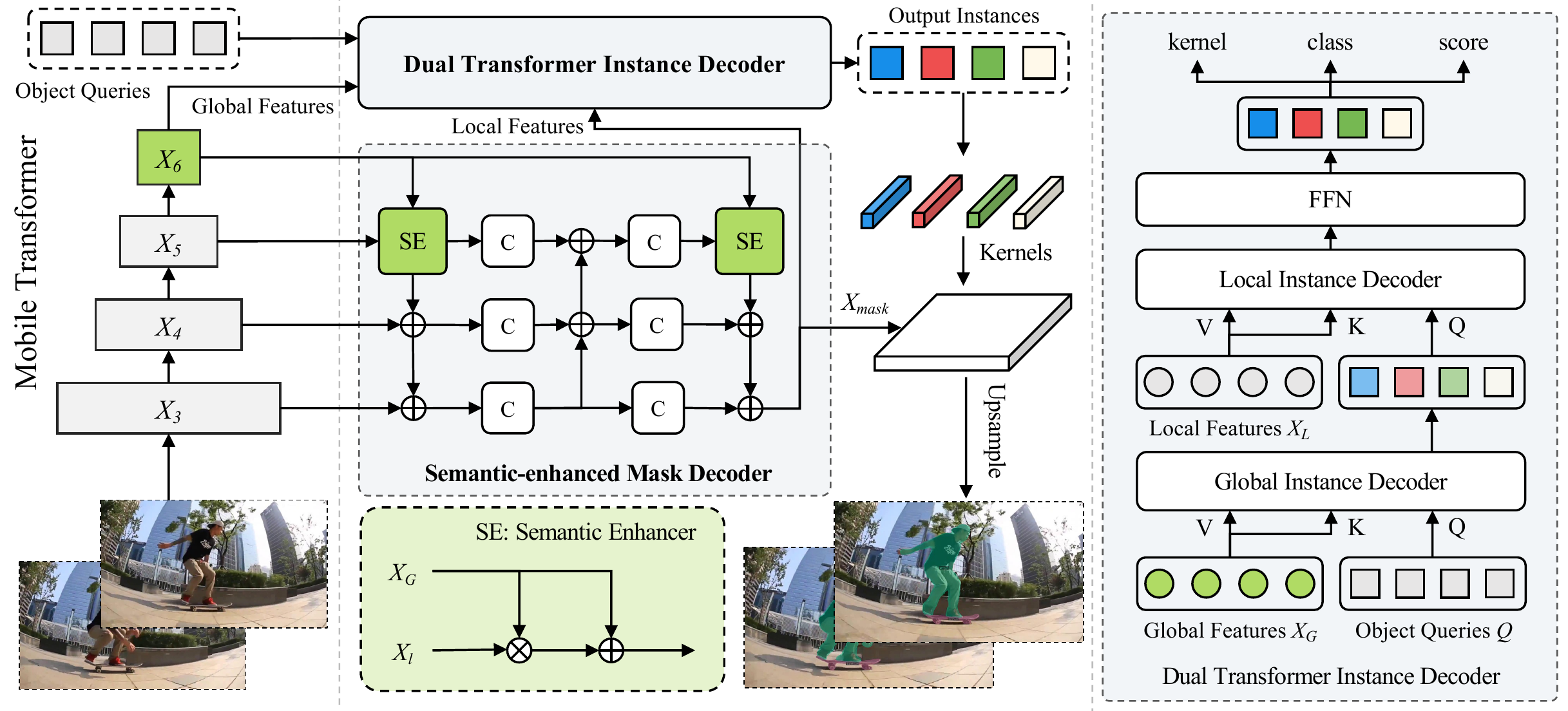}
    \vspace{-10pt}
    \caption{\textbf{Overall architecture of \name{}.} \name{} contains a mobile transformer as the backbone, a dual transformer instance decoder with learnable object queries to obtain object classes and kernels (Sec.~\ref{sec:inst_decoder}), and a semantic-enhanced mask decoder to obtain single-level features of high-semantics (Sec.~\ref{sec:mask_decoder}) by the semantic enhancers with global features $X_G$ ($X_6$ from the mobile transformer).
    The generated kernels from instance queries and mask features $X_{mask}$ can directly output the instance masks through the dot product. `C' in the square denotes $3\times3$ convolution.}
    \label{fig:main_arch}
    \vspace{-15pt}
\end{figure*}

%% file: tex/method.tex
\subsection{Overall Architecture}
\label{sec:overall_arch}
We present \name{}, a video instance segmentation framework tailor-made for mobile devices.
Fig.~\ref{fig:main_arch} gives an illustration of our framework.
Given input images, \name{} firstly utilizes a mobile transformer backbone to extract multi-level pyramid features.
Following \cite{TopFormer,SeaFormer}, our backbone network consists of a series of convolutional blocks and transformer blocks. It takes images as inputs and generates both local features (\ie., $X_3$, $X_4$, and $X_5$ in Fig.~\ref{fig:main_arch}) and global features (\ie, $X_6$).

Considering the global features $X_6$ contain abundant high-level semantic information, we present (1) \textit{dual transformer instance decoder} which adopts a query-based transformer decoder based on the global image features and local image features and generates the instance predictions, \ie, instance kernels and classification scores; (2) \textit{semantic-enhanced mask decoder} which employs the multi-scale features from the backbone and a \textit{semantic enhancer} to enrich the multi-scale features with semantic information.

\subsection{Dual Transformer Instance Decoder}
\label{sec:inst_decoder}
\paragraph{Queries are good trackers.} Detection transformers with object queries~\cite{DETRCarionMSUKZ20,DeDETRZhuSLLWD21} have achieved great progress in object detection.
Compared to previous dense detectors~\cite{faster,retinanet,fcos}, query-based detectors generate a sparse set of predictions and get rid of heuristic post-processing for duplicate removal.
Furthermore, previous methods~\cite{VIS,Crossvis} extend dense detectors for video instance segmentation by designing heuristic matching to associate instances across frames, which is inefficient and hard to optimize in mobile devices.
Whereas, as shown in Fig.~\ref{fig:kernel_reuse}, \textit{object queries are good trackers} and can be used to associate objects in videos based on three reasons:
(1) object queries are trained to segment the foreground of corresponding visual instance, thus naturally comprising contextualized instance features;
(2) object queries are forced to match objects in a one-to-one manner and duplicate queries are suppressed;
(3) the object query tends to be temporally consistent and represents the same instance in consecutive frames, which can be attributed to the temporal smoothness in adjacent frames.
Therefore, using object queries as trackers can omit complex heuristics post-process for associating objects and is more efficient on mobile devices.

However, directly attaching transformer decoders like \cite{DETRCarionMSUKZ20} on the mobile backbone leads to unaffordable computation budgets for mobile devices, and simply reducing decoder layers or parameters leads to unsatisfactory performance.
Striking the balance and designing mobile-friendly architectures is non-trivial and critical for real-world applications.
For efficiency, we present \textit{dual transformer instance encoder}, which simplifies the prevalent 6-stage decoders in \cite{DETRCarionMSUKZ20,DeDETRZhuSLLWD21} into 2-stage dual decoders, \ie, the global instance decoder and the local instance decoder, which takes the global features $X_G$ and local features $X_L$ as key and value for updating object queries.
We follow ~\cite{mask2former} and adopt the sine position embedding for both global and local features.
The object queries $Q$ are learnable and random initialized.
\paragraph{Global and Local Instance Decoder.}
Adding transformer encoders~\cite{DETRCarionMSUKZ20,DeDETRZhuSLLWD21} for the global contexts will incur a significant computation burden.
Instead, we adopt high-level features ($X_6$ in Fig.~\ref{fig:main_arch}) $X_6$ as global features $X_G$ for query update, which contains high-level semantics and coarse localization.
Inspired by recent works~\cite{mask_former_abs-2107-06278}, we adopt the fine-grained local features, \ie, the mask features $X_{mask}$, to compensate for spatial details for generating mask kernels.
For efficiency, we downsample the mask features to $\frac{1}{64}\times$ through \textit{max pooling}, \ie, $X_L = f_{\text{pool}}(X_{mask})$, which can preserve more details.
The \textit{dual transformer instance decoder} acquires contextual features from the global features $X_G$ and refines queries with fine-grained local features $X_L$.

\subsection{Semantic-enhanced Mask Decoder}
\label{sec:mask_decoder}
Multi-scale features are important for instance segmentation due to the severe scale variation in natural scenes.
In addition, generating masks requires high-resolution features for accurate localization and segmentation quality.
To this end, prevalent methods~\cite{mask_former_abs-2107-06278,mask2former} stack multi-scale transformers~\cite{mask2former} as pixel decoders to enhance the multi-scale representation and generate high-resolution mask features.
Stacking transformers for high-resolution features leads to large computation and memory costs.
Instead of using transformers, \cite{ChengWCZZHZ022} presents a FPN-PPM encoder with 4 consecutive $3\times3$ convolutions as mask decoder, which also leads to a huge burden, \ie, 7.6 GFLOPs.
For mobile devices, we thus present an efficient \textit{semantic-enhanced mask decoder}, as shown in Fig.~\ref{fig:main_arch}.
The mask decoder adopts the multi-scale features $\{X_3, X_4, X_5\}$ and outputs single-level high-resolution mask features ($\frac{1}{8}\times$). 
Motivated by FPN~\cite{LinDGHHB17}, we use iterative \textit{top-down} and \textit{bottom-up} multi-scale fusion.
Furthermore, we present the \textit{semantic enhancers} to strengthen the contextual information for the mask features with the global features $X_6$, as shown in the green blocks of Fig.~\ref{fig:main_arch}. Then the mask features $X_{mask}$ and the generated kernels $K$ are fused by $M=K \cdot X_{mask}$ to obtain the output segmentation masks.

\subsection{Tracking with Kernel Reuse and Association}
\label{sec:kernel_reuse}
As discussed in Sec.~\ref{sec:inst_decoder}, mask kernels (generated by object queries) are temporally consistent due to the temporal smoothness in adjacent frames. 
Hence, mask kernels can be directly adopted to segment and track the same instance in the nearby frames, \eg, 11 frames as shown in Fig.~\ref{fig:kernel_reuse}.
We thus present the efficient \textit{kernel reuse} to adopt the mask kernels from the keyframe to generate the segmentation masks for the consecutive $T-1$ frames as follows:
\begin{equation}
\label{eq:kernel_fuse}
\begin{aligned}
    M^t &= K^t\cdot X^t_{mask}, \\
    M^{t+i} &= K^t\cdot X^{t+i}_{mask}, i \in(0,...,T-1), \\
\end{aligned}
\end{equation}
where $\{M^i\}^{T-1+t}_{i=t}$ are the segmentation masks for the same instance in the T-frame clip, and $K^t$ is the reused mask kernel.
Different from clip-based methods~\cite{vistr,heo2022vita,wu2022seqformer}, kernel reuse performs on-the-fly segmentation and tracking given per-frame input.

However, kernel reuse tends to fail in long-time sequences or frames with drastic changes.
To remedy these issues, we follow \cite{minvis} and present a simple yet effective \textit{kernel association}, which uses cosine similarity between the consecutive keyframes. 
Under one-to-one correspondence, duplicate queries (kernels) tend to be suppressed, which enables simple similarity metrics to associate kernels of consecutive keyframes.
Compared to previous methods~\cite{VIS,Crossvis} based on sophisticated metrics and post-processing methods, the proposed kernel association is much simple and easy to deploy on mobile devices.

\name{} can be straightforwardly extended to video instance segmentation by incorporating the presented kernel reuse and association.
And experimental results indicate that \name{} using $T=3$ 
 can achieve competitive performance, as discussed in Tab.~\ref{tab:TEMPORAL}.
For simpler videos or scenes, the reuse interval $T$ can be further extended for more efficient segmentation and tracking.

\subsection{Temporal Training via Query Passing}

\begin{figure}
    \centering
    \includegraphics[width=\linewidth]{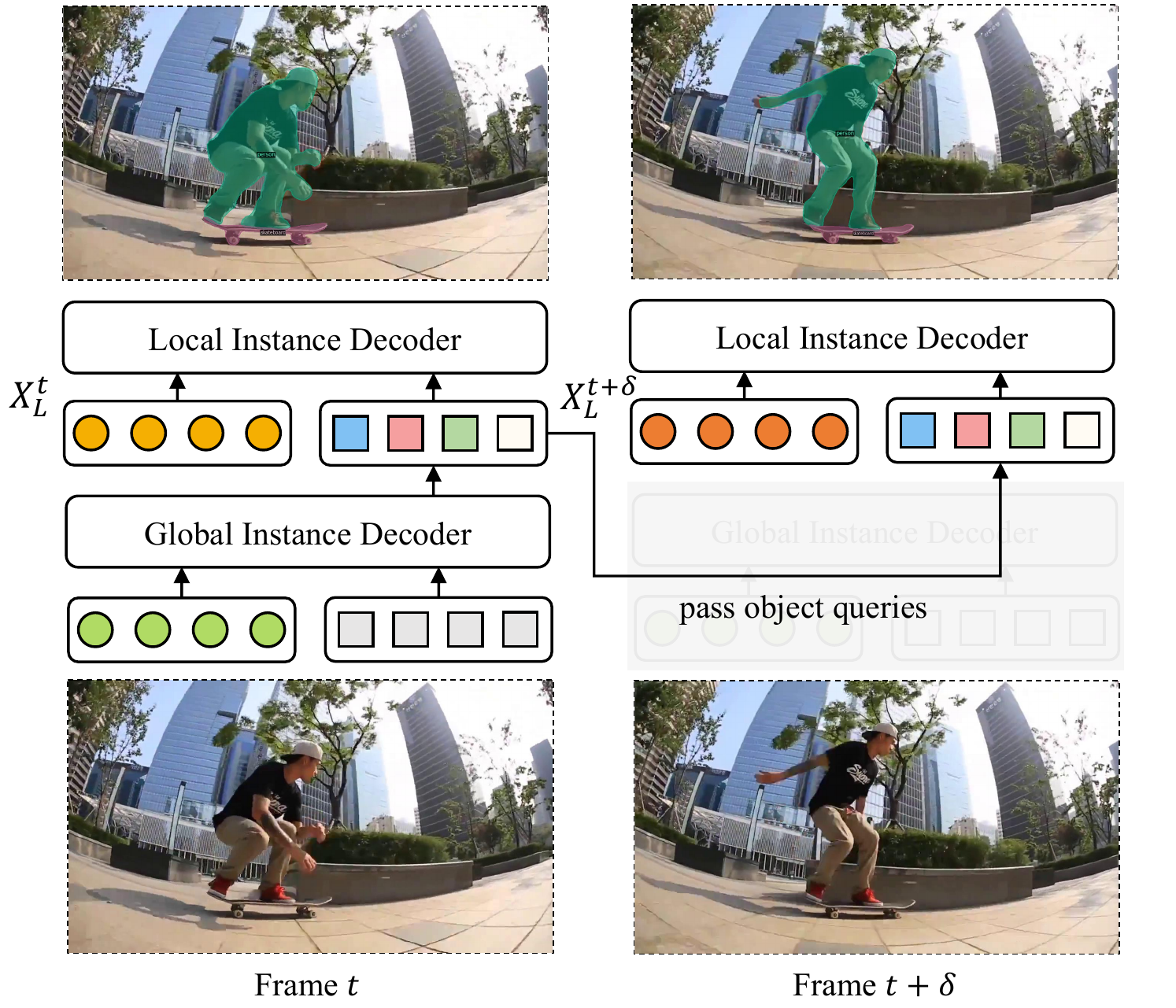}
    \vspace{-18pt}
    \caption{\textbf{Temporal Training via Query Passing.} We sample two frames with an interval $\delta$, \eg, the frame $t$ and the frame $t+\delta$. During temporal training, we adopt the object queries $Q_G^t$ of frame $t$ from the global instance decoder as the object queries $\tilde{Q}^{t+\delta}_G$ and pass it to the local instance decoder with local features $X^{t+\delta}_L$ to generate $\tilde{M}^{t+\delta}$.}
    \label{fig:query_passing}
    \vspace{-15pt}
\end{figure}

How to fully leverage temporal contextualized information in video for better temporal segmentation is a long-standing research problem in VIS.
Whereas, adding additional temporal modules introduces extra parameters and inevitably modifies the current architecture of \name{}.
To leverage temporal information in videos, we present a new temporal training strategy via query passing to enhance the feature representation for temporal inputs, which is inspired by~\cite{Crossvis}.
Specifically, we randomly sample two frames, \eg, frame $t$ and frame $t+\delta$, from a video sequence during training, as shown in Fig.~\ref{fig:query_passing}.
We adopt the object queries $Q^{t}_G$ generated from the global instance decoder as passing queries.
For frame $t+\delta$, we can obtain the mask features $X^{t+\delta}_{mask}$ and local features $X^{t+\delta}_L$ by normal forwarding.
During temporal training, the passing queries $Q^{t}_G$, as $\tilde{Q}^{t+\delta}_G$, are input to the local instance decoder with local features $X^{t+\delta}_L$ to obtain the kernels and generate masks $\tilde{M}^{t+\delta}$. 
The generated $\tilde{M}^{t+\delta}$ shares the same mask targets with $M^{t+\delta}$, and is supervised by the mask losses mentioned in Sec.~\ref{sec:loss}.

\label{sec:temporal_training}

\subsection{Loss Function}
\label{sec:loss}
\name{} generates $N$ predictions and uses bipartite matching for label assignment~\cite{DETRCarionMSUKZ20}. As the query passing does not require extra module and loss, we follow previous work~\cite{ChengWCZZHZ022} and use the same loss function for image/video instance segmentation, which is defined as follows:
\begin{equation}
\mathcal{L}=\lambda_c \cdot \mathcal{L}_{c l s}+\lambda_{mask} \cdot \mathcal{L}_{mask}+\lambda_{obj} \cdot \mathcal{L}_{obj},    
\end{equation}
where $\mathcal{L}_{cls}$ indicates the focal loss for classification, $\mathcal{L}_{mask}$ is the combination of dice loss and pixel-wise binary cross entropy loss for mask prediction, and $\mathcal{L}_{obj}$ indicates the binary cross-entropy loss for IoU-aware objectness. $\lambda_c$, $\lambda_{mask}$ and $\lambda_{obj}$ are set to 2.0, 2.0 and 1.0 respectively.

%% file: tex/exp.tex
In this section, we mainly evaluate \name{} on the challenging COCO~\cite{COCOLinMBHPRDZ14} and Youtube-VIS~\cite{VIS} datasets to demonstrate the effects of \name{} in terms of speed and accuracy\footnote{ALL Datasets were solely downloaded and evaluated by the University.}.
In addition, we conduct extensive ablation studies to reveal the effects of the components in \name{}.
We strongly refer the reader to the Appendix for additional experiments and visualizations.

\subsection{Datasets.}

\paragraph{COCO.}
COCO~\cite{COCOLinMBHPRDZ14} dataset is a touchstone for instance segmentation methods, which contains 118k, 5k, and 20k images for training, validation, and testing respectively.
\name{} is trained on \texttt{train2017} and evaluated on \texttt{val2017} or \texttt{test-dev2017}.

\paragraph{YouTube-VIS.}
YouTube-VIS 2019~\cite{VIS} is a large-scale dataset for video instance segmentation, which contains $2,883$ videos and $4,883$ instances covering $40$ object categories. As an extension, YouTube-VIS 2021 expands it to $1.5\times$ videos and $2\times$ instances with improved $40$ categories. Following previous works, we evaluate our methods on the validation set of both datasets.

\begin{table*}[t]
    \centering
    \renewcommand{\tabcolsep}{4pt}
    \renewcommand\arraystretch{1.1}
    \small
    \begin{tabular}{l|l|c|c|ccc|ccc}
    method & backbone & size &latency(ms) & \apcoco \\
    \tline
    Mask R-CNN~\cite{HeGDG17} & R-50 & 800 & - & 37.5  & 59.3 &  40.2  & 21.1  & 39.6  & 48.3\\
    CondInst~\cite{TianSC20} & R-50 & 800 & 4451 & 37.8 & 59.1&  40.5&  21.0&  40.3&  48.7\\
    SOLOv2-Lite~\cite{solov2} & R-18 & 448 & 693 & 30.0 & 48.6 & 
    31.4 & 8.4 &30.3 & 48.7\\
    SOLOv2-Lite~\cite{solov2} & R-50 & 448 & 1234 & 34.0 & 54.0 & 36.1 & 10.3 & 36.3 & 54.4 \\
    YOLACT~\cite{YolactBolyaZXL19} & R-50  & 550 & 1039 &28.2 & 46.6 & 
    29.2 & 9.2 &29.3 & 44.8\\
    SparseInst~\cite{ChengWCZZHZ022} & R-50 & 608 & 1752 &34.7 & 55.3 &  36.6 &  14.3 &  36.2 &  50.7 \\
    \hline
    YOLACT~\cite{YolactBolyaZXL19} & MobileNetV2 & 550 & 463 & 22.2&37.7&22.5&6.0&21.3&35.5 \\
    SOLOv2~\cite{solov2} & MobileNetV2& 640 & 1443 & 30.5 & 49.3& 32.1 & 4.2 & 49.6& 67.9   \\
    YOLACT~\cite{YolactBolyaZXL19} & TopFormer& 550 & 497 & 20.8&37.6&20.2&6.0&20.1&33.5 \\
    CondInst~\cite{TianSC20} & TopFormer & 640 & 1418 & 27.0 & 44.8 & 28.0 & 11.4 & 27.7& 39.0\\
    SparseInst~\cite{ChengWCZZHZ022} & TopFormer &608 & 769 & 30.0 & 49.2 & 30.9 & 11.0 & 29.5 & 46.2 \\
    Mask2Former$^{\dag}$~\cite{mask2former} & TopFormer &640 & 930 & 32.0 & 51.9 & 33.4 & 6.9 & 49.3 & 68.7 \\
    FastInst~\cite{fastinst} & TopFormer & 640 & 965 &  31.0 & 50.8 & 32.0 & 9.7 & 31.1 & 51.7 \\
    \hline
    \textbf{\name{}} & MobileNetV2 & 640 & 410 & 30.0&49.7&30.8&10.3&30.2&46.0\\
    \textbf{\name{}} & TopFormer & 512 & 346 & 29.9&49.4&30.6&9.0&29.2&48.5\\
    \textbf{\name{}} & TopFormer & 640 & 433 & 31.2&51.4&32.1&10.4&31.3&49.1\\
    \textbf{\name{}} & SeaFormer & 640 & 438 & 31.6&51.8&32.6&10.0&31.5&50.8\\
    \end{tabular}
    \vspace{-10pt}
    \caption{\textbf{Instance Segmentation on COCO \texttt{test-dev}.} Comparisons with state-of-the-art methods for mask AP and inference latency on COCO \texttt{test-dev}. The method denoted with $\dag$ was implemented by us.}
    \label{tab:coco_main_experiments}
    \vspace{-10pt}

\end{table*}

\begin{table*}[t]
    \centering
    \renewcommand{\tabcolsep}{4pt}
    \renewcommand\arraystretch{1.1}
    \small
    \begin{tabular}{l|l|cc|cccc|cccc}
   \multirow{2}{*}{method} & \multirow{2}{*}{backbone} & \multirow{2}{*}{\makecell[c]{GPU\\(ms)}} & \multirow{2}{*}{\makecell[c]{Mobile\\(ms)}}& \multicolumn{4}{c|}{YouTube-VIS 2019} & \multicolumn{4}{c}{YouTube-VIS 2021} \\
   \cline{5-12}
    & & & & \aps & AR$_{\text{1}}$& \aps & AR$_{\text{1}}$\\
    \tline
    Mask Track R-CNN~\cite{VIS} & R-50 & 30.1 & - & 30.3 & 51.1 & 32.6 &34 & 28.6 & 48.9 & 29.6 & 26.5 \\
    SipMask~\cite{SipMask} & R-50 & 29.3 & - & 33.7 & 54.1 & 35.8 & 35.4 & 31.7 & 52.5 & 34.0 & 30.8 \\
    SGMask~\cite{sgnet} & R-50 & 31.9 & - & 34.8 & 56.1 & 36.8 & 35.8 & - & - & - & - \\
    STMask~\cite{stmask} & R-50 & 28.2 & - & 33.5 & 52.1 & 36.9 & 31.1 & 30.6 & 49.4 & 32.0 & 26.4\\
    CrossVIS~\cite{Crossvis} & R-50 & 25.0 & 981 & 34.8 & 54.6 & 37.9 & 34.0 & 33.3 & 53.8 & 37.0 & 30.1\\
    \hline
    CrossVIS & TopFormer & 24.9 & 614 & 32.7 & 54.3 & 35.4 & 34 & 28.9 & 50.9 & 29.0 & 27.8\\
    SparseInst-VIS$^{\dag}$ & TopFormer & 25.4 &  389 & 33.3 & 55.1 & 34.1 &35.3 & 29.0 & 50.5 & 29.2 & 29.3\\
    \textbf{\name{}} & TopFormer  & 22.3 & 188 & 35.0 & 55.2 & 37.3 & 38.5 & 30.1 & 50.6 & 30.7 & 30.1\\
    \end{tabular}
    \vspace{-5pt}
    \caption{\textbf{Video Instance Segmentation on YouTube-VIS 2019 and YouTube-VIS 2021.} `GPU' denotes NVIDIA 2080 Ti and `Mobile' denotes Snapdragon 778G. The method denoted with $^{\dag}$ was implemented by us.}
    \label{tab:yvis_main_experiments_19_21}
    \vspace{-15pt}

\end{table*}

\subsection{Implementation Details}

\paragraph{Instance Segmentation.}
We use the AdamW optimizer with an initial learning rate $1 \times 10^{-4}$ and set the backbone multiplier to $0.5$. Following the training schedule and data augmentation as \cite{ChengWCZZHZ022}, all models are trained for $270k$ iterations with a batch size of $64$ and decay the learning rate by $10$ at $210k$ and $250k$. We apply random flip and scale jitter to augment the training images. More precisely, the shorter edge varies from $416$ to $640$ pixels, while the longer edge remains under $864$ pixels.

\paragraph{Video Instance Segmentation.}
The models are initialized with weights from the instance segmentation model pre-trained on the COCO \texttt{train2017} set. We set the learning rate of instance segmentation to $5 \times 10^{-5}$ and train for $12$ epochs with a $10\times$ decay for learning rate at the $8$th and $11$th epochs. We only employ basic data augmentation, such as resizing the shorter side of the image to $360$, without using any additional data or tricks.

\paragraph{Inference.}

The inference of \name{} is simple.
\name{} can directly output the instance segmentation results for single-frame images without non-maximum suppression (NMS).
The inference speeds of all models are measured using TNN framework\footnote{TNN: a uniform deep learning inference framework} on the CPU core of Snapdragon 778G without other methods of acceleration.

\subsection{Experiments on Instance Segmentation}
Firstly, we evaluate the proposed \name{} on COCO \texttt{test-dev} dataset for mobile instance segmentation. As the first instance segmentation model designed specifically for mobile devices, we benchmark our approach against real-time instance segmentation methods. Tab.~\ref{tab:coco_main_experiments} shows the comparisons between \name{} and previous approaches.

Among all the methods which use ResNet~\cite{HeZRS16} backbone, Mask R-CNN and CondInst naturally achieve AP above $37$. However, the deployment challenges of Mask R-CNN as a two-stage model and CondInst make them less desirable for mobile applications. We observe that \name{} achieves higher accuracy than the popular real-time approach YOLACT based on R-50, with an increase of $3.4$ AP and $600$ ms faster speed. Notably, \name{} obtains faster inference speed and higher accuracy compared to those methods~\cite{YolactBolyaZXL19,SOLOV2WangZKLS20,TianSC20,ChengWCZZHZ022,mask2former} with lightweight backbones~\cite{Sandler2018MobileNetV2IR,TopFormer}.
Tab.~\ref{tab:coco_main_experiments} shows a remarkable speed improvement of up to $50\%$ compared to the previous state-of-the-art method SparseInst. Compared to the well-established Mask2Former, \name{} has a similar AP with $100\%$ speed improvement. Fig.~\ref{fig:speed_and_accuracy} illustrates the trade-off curve between speed and accuracy, which further clearly shows the great performance of \name{}.

\subsection{Experiments on Video Instance Segmentation}
In Tab.~\ref{tab:yvis_main_experiments_19_21}, we evaluate \name{} YouTube-VIS 2019 and YouTube-VIS 2021 for video instance segmentation.
In terms of latency and accuracy, we mainly compared \name{} with online methods. As shown In Tab.~\ref{tab:yvis_main_experiments_19_21}, \name{} can obtain better accuracy and speed than \cite{Crossvis,ChengWCZZHZ022} under the same setting.
Considering that TopFormer aims for mobile devices and it's less efficient on GPU. 
However, it is still evident that \name{} has superior inference speed on mobile devices.

\subsection{Ablation Study}
\paragraph{Ablation on Instance Decoder.} In Tab.~\ref{tab:ablation_decoder_global_local}, We evaluate the performance and speed of different configurations of the instance decoder. Tab.~\ref{tab:ablation_decoder_global_local} shows that using a single global instance decoder or a single local instance decoder leads to a performance drop, which demonstrates the effectiveness of the instance decoder with global features for semantic contexts and local features for spatial details. 
Stacking two local instance decoders obtains a similar performance with the proposed instance decoder, \ie, $29.8$ mask AP. However, Tab.~\ref{tab:TEMPORAL} indicates that the \textit{proposed instance decoder with aggregating global contexts is superior to stacking two local decoders in terms of segmenting and tracking} in videos.

In Tab.~\ref{tab:ablation_pool}, we mainly focus on the local instance decoder and compare  different methods of extracting local features from mask features: no pooling, max pooling with the kernel size of $4$ or $8$, and average pooling with a kernel size of $8$. Although no pooling provides a gain of $0.9$ in AP, it also incurs a $50\%$ increase in latency, making it not cost-effective. Additionally, it is worth noting that using max pooling leads to a $0.4$ AP gain compared to using average pooling. We believe max pooling naturally provides more desirable local information by filtering out unimportant information, forming a better complementary relationship with the global features used in the global instance decoder.

\begin{table}
    \centering
    \renewcommand{\tabcolsep}{4pt}
    \renewcommand\arraystretch{1.1}
    \small
    \begin{tabular}{cc|ccc|cc}
    global & local & \apscoco & latency & FLOPs \\
    \tline
    \checkmark & &28.7 &46.9 & 29.5 &413ms &24.17G \\
     & \checkmark &29.3 &48.9 &30.2 & \tb{412ms} & \tb{24.15G} \\
    \hline
    
    $\times2$ & & 28.5& 46.3 & 29.5 &427ms &24.24G \\
     & $\times2$ & \tb{29.8} & 49.3 & \tb{30.4} & 426ms &24.22G \\
    \checkmark & \checkmark & \tb{29.8} & \tb{49.4} & \tb{30.4} & 427ms & 24.24G \\
    \end{tabular}
    \vspace{-5pt}
    \caption{\textbf{Ablation on the Instance Decoder (COCO \texttt{val2017}).} Both the global decoder and local decoder contribute to improvement. $\times2$ indicates stacking two decoders. Despite the similar performance, global-local is better than local-local for VIS tasks (refer to Tab.~\ref{tab:TEMPORAL}).}
    \label{tab:ablation_decoder_global_local}
    \vspace{-10pt}
\end{table}

\begin{table}
    \centering
    \renewcommand{\tabcolsep}{4pt}
    \renewcommand\arraystretch{1.1}
    \small
    \begin{tabular}{l|c|ccc|ccc}
    \multirow{2}{*}{decoder} & \multirow{2}{*}{AP$^{\text{COCO}}$}& 
    \multicolumn{3}{c|}{w/o tem. training} & \multicolumn{3}{c}{w/ tem. training} \\
    \cline{3-8}
    & & T=1 & T=3 & T=6 & T=1 & T=3 & T=6 \\

    \tline
    global-local & 29.8 & 28.8 & \tb{29.3}  &28.0  & 30.1 & \tb{30.5} & 29.2  \\
    local-local  & 29.8 & 28.1 & \tb{28.5} & 27.8  & 28.9 & \tb{29.5} & 28.6 \\
    \hline
    latency (ms) & 184 & 184 & 174  & 171 & 184 & 174 & 171 \\
    \end{tabular}
    \vspace{-5pt}
    \caption{\textbf{Ablation on the Query Reuse \& Temporal Training (YouTube-VIS 2021).} `T' refers to the length of the clip within which we reuse mask kernels of the keyframe. 
    Single-frame clips ($T=1$) only associate kernels without reuse. 
    }
    \label{tab:TEMPORAL}
    \vspace{-10pt}
\end{table}

\paragraph{Kernel Reuse \& Temporal Training.}
We conduct a comparative study of two decoder designs (refer to Tab.~\ref{tab:ablation_decoder_global_local}), \ie, (1) global-local: the combination of a global instance decoder and a local instance decoder and (2) local-local: two local instance decoders, as shown in Tab.~\ref{tab:TEMPORAL}.
For kernel Reuse, $T$ refers to the length of the clip within which we reuse the mask kernels of the keyframe. Regardless of the model architecture, the reuse mechanism in short-term sequences improves inference speed without performance loss. 
Compared to the training with only frame-level information, the proposed temporal training brings $1.3$ and $0.8$ AP improvement for the two designs, respectively. 
In terms of the global-local and local-local decoders, Tab.~\ref{tab:TEMPORAL} shows that global-local achieves better performance on video instance segmentation. 
Compared to the local-local decoder, the queries (kernels) from the global-local decoder aggregate more global contextual features and benefits more from temporal smoothness in videos, as discussed in Sec.~\ref{sec:inst_decoder}, which is more suitable for videos.
Tab.~\ref{tab:TEMPORAL} well demonstrates the proposed dual transformer instance decoder for video instance segmentation.

\begin{table}
    \centering
    \renewcommand{\tabcolsep}{6pt}
    \renewcommand\arraystretch{1.1}
    \small
    \begin{tabular}{cc|ccc|cc}
    size & pool & \apscoco & latency & FLOPs \\
    \tline
    ori. & - & \tb{30.7} & \tb{51.1} & \tb{31.3} & 613ms & 25.85G\\
    $4\!\times\!4$ & max & 30.0 &50.2 & 30.8 &434ms& 24.32G \\
    $8\!\times\!8$ & max &29.8 &49.4 &30.4 &\tb{427ms}& \tb{24.24G}\\
    $8\!\times\!8$ & avg & 29.4 & 48.4 &30.3 & \tb{427ms} & \tb{24.24G} \\
    \end{tabular}
    \vspace{-5pt}
    \caption{\textbf{Ablation on the Local Instance Decoder (COCO \texttt{val2017}).} 
    The pooling is used to extract local features from the mask features for the local instance decoder. Decreasing the pool size can further improve the accuracy but lower the speed. Notably, max pooling brings 0.4 AP gain compared to the average pooling.}
    \label{tab:ablation_pool}
    \vspace{-10pt}
\end{table}

\paragraph{Ablation on the Mask Decoder.}
Mask features play a crucial role in segmentation quality.  Here, we investigate different designs of mask decoders in Tab \ref{tab:ablation_mask_decoder}. Compared to FPN with $1\times$ conv, our method achieves $1.1$ AP improvement by utilizing multi-scale information in an iterative manner, with a latency overhead of only 6ms. Although stacking convolutions still improves the performance, as seen from the results of SparseInst with $4$ stacked $3\times 3$ convs, it leads to a significant burden for mobile devices. The proposed semantic-enhancer (SE) brings $0.3$ AP improvement, and bridges the gap with less cost.

\begin{table}
    \centering
    \renewcommand{\tabcolsep}{4pt}
    \renewcommand\arraystretch{1.1}
    \small
    \begin{tabular}{l|ccc|cc}
    mask decoder & \apscoco & latency & FLOPs \\
    \tline
    SparseInst, $4\times$conv & \tb{30.4} & 49.6 & \tb{31.2} & 524ms&34.69G \\
    SparseInst, $2\times$conv & 29.7 & 49.2 & 30.4 &445ms &24.11G\\
    SparseInst, $1\times$conv &29.1 &
    48.8 &29.7  & 405ms &18.82G \\
    \hline
    FPN, $1\times$conv & 28.7 & 48.1 & 29.2 & \tb{400ms} & \tb{18.48G} \\
    ours &29.8 & 49.4 & 30.4 &427ms & 24.24G \\
    ours w/ SE  & 30.1 & \tb{49.9} & 30.9 & 433ms &24.37G \\
    \end{tabular}
    \vspace{-5pt}
    \caption{\textbf{Ablation on the Semantic-enriched Decoder (COCO \texttt{val2017}).} `SparseInst' denotes the FPN-PPM used in~\cite{ChengWCZZHZ022}. Our mask decoder obtains $1.4$ AP improvement compared to the FPN, and semantic-enhancer SE) brings $0.3$ AP gain.}
    \label{tab:ablation_mask_decoder}
    \vspace{-15pt}
\end{table}

%% file: tex/vis.tex
In this section, we provide qualitative results on both the COCO dataset and the YouTube-VIS dataset.
\subsection{Instance Segmentation}
\begin{figure*}[t]
    \centering
    \includegraphics[width=\linewidth]{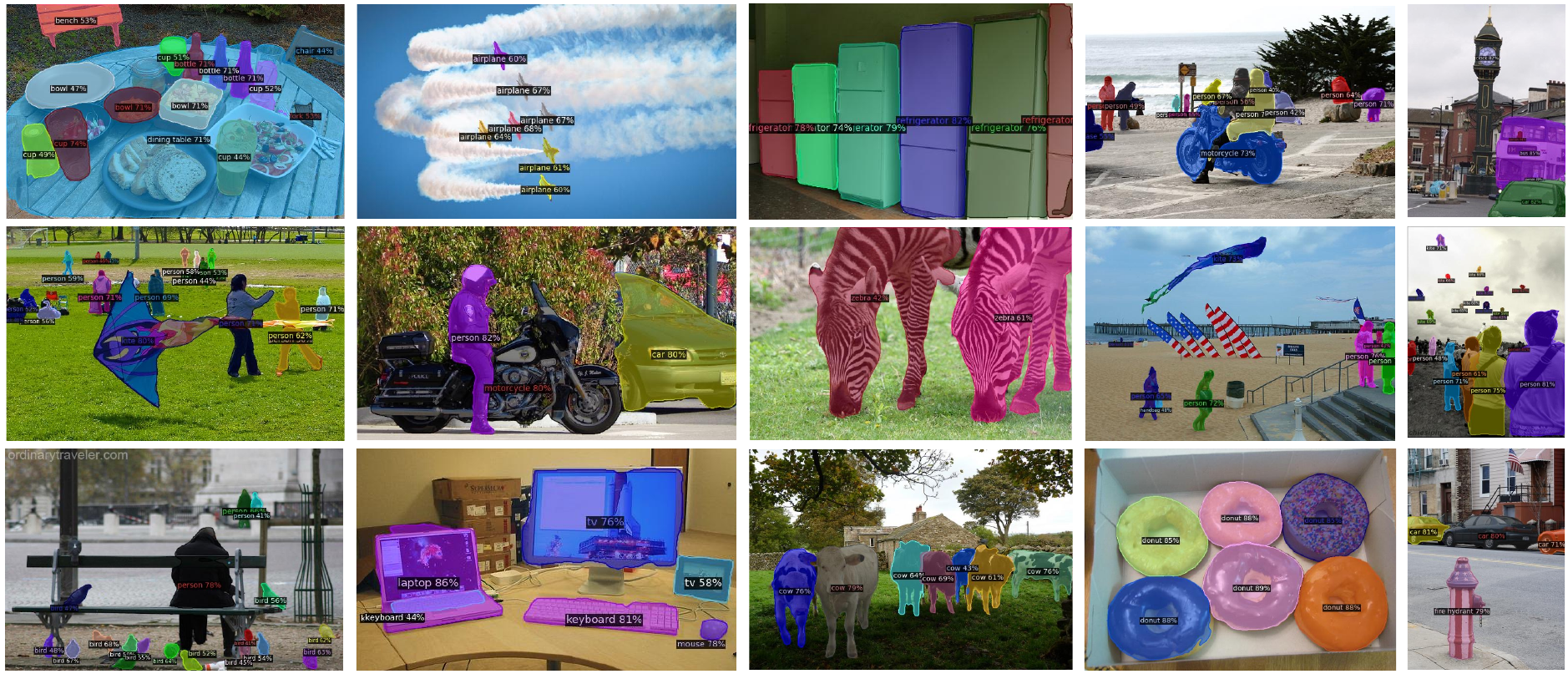}
    \caption{\textbf{Qualitative Results on COCO for Instance Segmentation.} MobileInst with TopFormer as backbone achieves 30.1 mAP on COCO \texttt{val2017}. The confidence threshold is set to 0.4.}
    \label{fig:coco_vis}
\end{figure*}

Fig.~\ref{fig:coco_vis} shows the qualitative results of \name{} on COCO for instance segmentation.
\name{} can well segment objects with fine boundaries.

\subsection{Video Instance Segmentation}

\begin{figure*}[t]
    \centering
    \includegraphics[width=\linewidth]{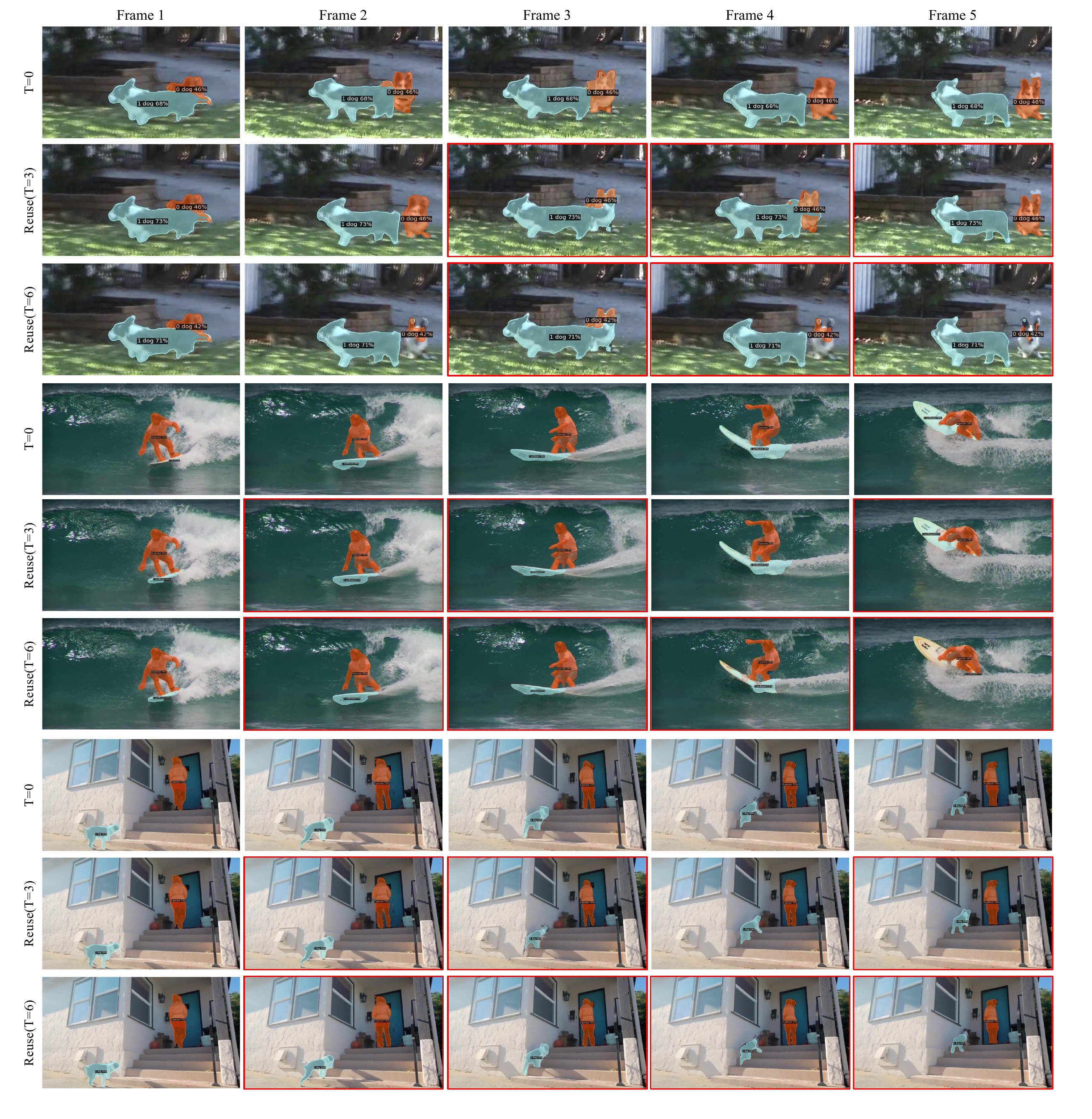}
    \caption{\textbf{Qualitative Results on YouTube-VIS-2021 for Video Instance Segmentation (A).} `T' refers to the length of the clip within which we reuse mask kernels of the keyframe.  
    Frames with \textcolor{red}{red rectangles} are non-keyframes and reuse the mask kernel and classification score from the last keyframe.
    The motion of instances might lead to coarse boundaries of the segmentation.
    }
    \label{fig:yvis_vis_1}
\end{figure*}

\begin{figure*}[t]
    \centering
    \includegraphics[width=\linewidth]{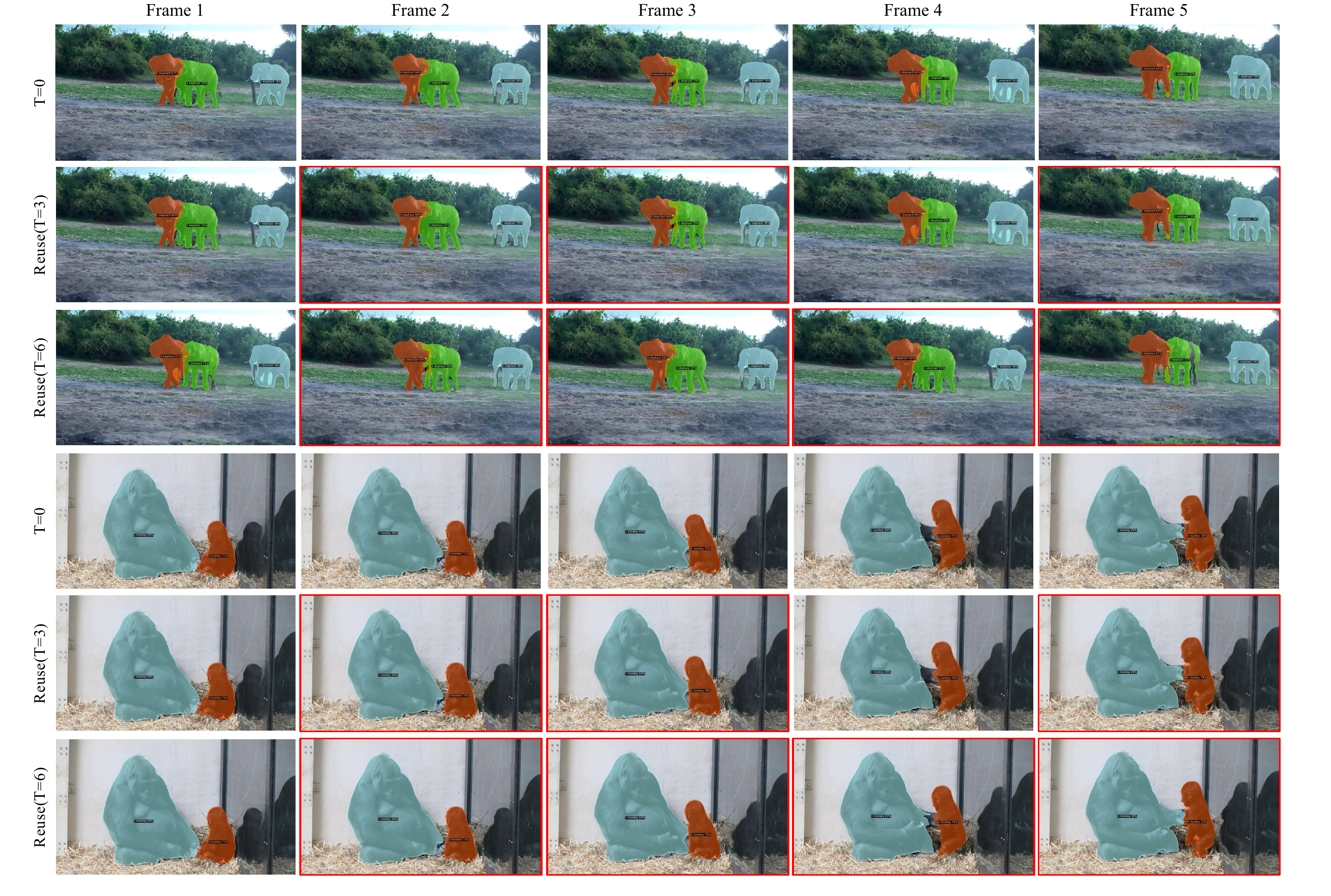}
    \caption{\textbf{Qualitative Results on YouTube-VIS-2021 for Video Instance Segmentation (B).} `T' refers to the length of the clip within which we reuse mask kernels of the keyframe. Frames with \textcolor{red}{red rectangles} are non-keyframes and reuse the mask kernel and classification score from the last keyframe.}
    \label{fig:yvis_vis_2}
\end{figure*}

Fig.~\ref{fig:yvis_vis_1} and Fig.~\ref{fig:yvis_vis_2} show the qualitative results of \name{} for video instance segmentation on the YouTubeVIS-2021 validation set. It's clear that \name{} can effectively segment and track instances, thus achieving impressive results.
Under some complex scenes with severe motion, as shown in Fig.~\ref{fig:yvis_vis_1}, \name{} might suffer from the coarse boundaries while still obtaining good tracking results.
Whereas, \name{} performs well under the scenes without severe motion, as shown in Fig.~\ref{fig:yvis_vis_2}.

